\documentclass[10pt,twocolumn,letterpaper]{article}

\usepackage{iccv}
\usepackage{times}
\usepackage{epsfig}
\usepackage{graphicx}
\usepackage{amsmath}
\usepackage{amssymb}

\usepackage{adjustbox} 
\usepackage{multirow}
\usepackage{booktabs}

\usepackage[accsupp]{axessibility} 

\AddToShipoutPicture*{\footnotesize \sffamily\raisebox{1.6cm}[0pt][0pt]{
    {\hspace{1.4cm}
    \raisebox{0cm}{© 2021 IEEE. Personal use of this material is permitted. Permission from IEEE must be
    obtained for all other uses, in any current or future media,}%
    \hspace{-18.1cm}
    \raisebox{-0.4cm}{including
   reprinting/republishing this material for advertising or promotional purposes, creating new collective works, for resale or redistribution to}
    \hspace{-18.1cm}
    \raisebox{-0.8cm}{servers or lists, or reuse of any copyrighted component of this work in other works.}%
  }
} }

\usepackage[breaklinks=true,bookmarks=false]{hyperref}

\iccvfinalcopy 

\def\iccvPaperID{13} 

\ificcvfinal\fi
\begin{document}

\title{Effect of Parameter Optimization on Classical and Learning-based Image Matching Methods}

\author{Ufuk Efe, Kutalmis Gokalp Ince, A. Aydin Alatan\\
Department of Electrical and Electronics Engineering, Center for Image Analysis (OGAM)\\
Middle East Technical University, Ankara, Turkey\\
{\tt\small ufuk.efe, kutalmis, alatan @ metu.edu.tr}

}

\maketitle
\thispagestyle{empty}

\begin{abstract}
Deep learning-based image matching methods are improved significantly during the recent years. Although these methods are reported to outperform the classical techniques, the performance of the classical methods is not examined in detail. In this study, we compare classical and learning-based methods by employing mutual nearest neighbor search with ratio test and optimizing the ratio test threshold to achieve the best performance on two different performance metrics. After a fair comparison, the experimental results on HPatches dataset reveal that the performance gap between classical and learning-based methods is not that significant. Throughout the experiments, we demonstrated that SuperGlue is the state-of-the-art technique for the image matching problem on HPatches dataset. However, if a single parameter, namely ratio test threshold, is carefully optimized, a well-known traditional method SIFT performs quite close to SuperGlue and even outperforms in terms of mean matching accuracy (MMA) under 1 and 2 pixel thresholds. Moreover, a recent approach, DFM, which only uses pre-trained VGG features as descriptors and ratio test, is shown to outperform most of the well-trained learning-based methods. Therefore, we conclude that the parameters of any classical method should be  analyzed carefully before comparing against a learning-based technique.
\end{abstract}

\section{Introduction}

Determining pixel-to-pixel correspondences between two images is one of the fundamental problems in computer vision. There exists various classical "hand-crafted" approaches, such as SIFT \cite{lowe2004sift}, SURF \cite{bay2006surf}, ORB \cite{rublee2011orb}, KAZE \cite{alcantarilla2012kaze}, AKAZE \cite{alcantarilla2011akaze}, as well as some recently developed learning-based methods, SuperPoint \cite{detone2018superpoint}, SuperGlue \cite{sarlin2020superglue}, Patch2Pix \cite{zhou2020patch2pix} and DFM \cite{Efe_2021_CVPR}. All these techniques are frequently used in many applications, such as image matching, camera relocalization, pose estimation, Simultaneous Localization and Mapping (SLAM), and Structure-from-Motion (SfM).

Testing the performance of an algorithm in a fair manner for different applications is a challenging task. In addition, comparing and evaluating these algorithms on well-known datasets is also not straightforward due to hyper-parameter selection. This step is known to be crucial specifically for classical algorithms due to numerous parameters of such algorithms. In case of a comparison  between a new learning-based method to the classical methods, this parameter adjustment procedure is not examined in detail. Moreover, most of the time, the hyper-parameters of the classical methods are not even specified in the manuscripts. In this paper, we tried to compare classical and learning-based image matching algorithms in a fair manner on a well-known (HPatches) dataset \cite{balntas2017hpatches} by optimizing the hyper-parameters of each algorithm on the selected dataset.

In a recent study \cite{Efe_2021_CVPR}, we demonstrated that applying mutual nearest neighbor search that exploits the ratio test on pre-trained VGG \cite{simonyan2014vgg} features achieves the state-of-the-art performance. In order to investigate the resulting effect of this ratio threshold on some classical methods, we have also examined five classical image matching algorithms \cite{lowe2004sift, bay2006surf, rublee2011orb, alcantarilla2012kaze, alcantarilla2011akaze} on HPatches dataset in terms of \textit{Mean Matching Accuracy (MMA)} \cite{dusmanu2019d2net} and \textit{Homography Estimation Accuracy (HEA)} \cite{detone2018superpoint}. We have compared these conventional algorithms against four popular learning-based \cite{detone2018superpoint, sarlin2020superglue, zhou2020patch2pix, Efe_2021_CVPR} methods. We observed that ratio test threshold have a significant impact, as in \cite{jin2021imagematchingfrompapertopractice, dusmanu2020multiviewoptimization}, on the performance of the methods SIFT, SURF, ORB, KAZE, AKAZE, SuperPoint and DFM. Similarly, the confidence parameter utilized in SuperGlue and Patch2Pix algorithms also affects their performance. This study presents the result of comprehensive experiments on these nine algorithms with different ratio test and confidence thresholds to reveal MMA and HEA performances of those algorithms for 1-10 pixel thresholds. By using these results, we present the optimal parameters for each algorithm to maximize MMA and HEA for required pixel threshold.

Most of the new algorithm proposals claim to outperform the preceding ones at least for some specific pixel thresholds. As we present through the experiments, most of the time by adjusting the hyper-parameters, it is possible one algorithm to outperform the rest for a specific pixel threshold, resulting in multiple state-of-the-art methods. To minimize this ambiguity, beyond the pixel threshold specific optimal parameters, Area Under Curve (AUC) values for MMA and HEA using the whole accuracy range (1-10 pixel thresholds) are also provided  to reveal the average performance. We also present optimal parameters for MMA and HEA individually, since the optimal parameters for these two metrics might be different as demonstrated in \cite{zhou2020patch2pix, Efe_2021_CVPR}.

Throughout the experiments, we observed that changing only a single parameter of an algorithm yields this algorithm reaching the state-of-the-art performance. This circumstance is mostly observed for classical methods, which usually argued to have inferior performance compared to the learning-based algorithms. In other words, we demonstrate and argue that classical algorithms can still perform close to the state-of-the-art for image matching task, at least in HPatches dataset, and the performance gap between the classical and deep learning-based methods is not that significant, as it is accepted and reported before in the  literature.

Hence, in this study, our contribution is threefold;

i.	We show that, by only adjusting a single hyper-parameter, namely ratio test threshold, classical algorithms still competes with the state-of-the-art learning-based methods in terms of MMA and HEA metrics on HPatches dataset.

ii.	By revisiting the existing methods, we propose the optimal parameter settings for classical and learning-based algorithms and present the optimal MMA and HEA performance on HPatches dataset for each method.

iii.	We provide an experimental setup on \url{https://github.com/ufukefe/IME} to determine the optimal parameters of 9 well-known image matching algorithms using MMA and HEA metrics not only for HPatches but also for any dataset in which these performance metrics can be employed. 

\section{Related Work}
In order to determine pixel-wise correspondences between two images, the classical approaches follow detection, description, and matching steps, while learning-based methods either follow these steps or carry out all of these steps in a single stage. 

\subsection{Classical Image Matching Algorithms}
The most well known classical algorithms SIFT, SURF, ORB, KAZE and AKAZE execute the first two steps, namely feature detection and description, in a hand-crafted manner, and give locations of detected points together with their corresponding descriptors.

SIFT (Scale-Invariant Feature Transform) \cite{lowe2004sift} detects features in the scale-space simply utilizing Difference of Gaussians (DoG), which is an approximation of Laplacian of Gaussian (LoG) \cite{lindeberg1994scale}. In this manner, SIFT is able to detect blobs in varying scales with their orientation. Next, considering the dominant orientation and creating gradient histograms, SIFT outputs 128-dimensional descriptor vectors. SIFT is invariant to some amount of scale changes and even severe rotations; furthermore, it is robust to illumination changes due to the normalization of the descriptor vector.

SURF (Speeded Up Robust Features) \cite{bay2006surf} goes a little further than the SIFT and approximates LoG with low-complexity Box Filters. SURF uses a blob detector based on the Hessian matrix to find points of interest. The determinant of the Hessian matrix is used as a measure of local change around the point, and as a consequence, the points which maximize that determinant are selected. SURF also takes into account the dominant orientation of the features and exploits Haar wavelet responses in the horizontal and vertical directions to extract 64-dimensional descriptor vectors. SURF is also robust to some amount of scale, rotation, and illumination changes. 

ORB (Oriented FAST and Rotated BRIEF) \cite{rublee2011orb} runs FAST (Features from Accelerated Segment Test) \cite{rosten2005fast} as a feature point detector together with computing keypoint’s orientation. FAST basically examines a circle of 16 pixels surrounding an arbitrary pixel and decides whether the pixel is a keypoint or not by using the intensity differences between the candidate pixel and 16 surrounding pixels. Then, using the computed orientation, ORB employs a rotated version of the BRIEF (Binary Robust Independent Elementary Features) \cite{calonder2010brief} algorithm, which simply creates a binary feature vector of the binary test responses to build descriptor vectors. 

KAZE \cite{alcantarilla2012kaze} detector uses nonlinear scale spaces instead of Gaussian scale-space representations that are employed by SIFT. The motivation is that; nonlinear scale-space considers objects' natural boundaries, unlike Gaussian scale-space, which does not regard them due to smoothing the details and noise at all scale levels to the same degree. In contrast to SIFT, which uses the (DoG) to process the blurred images, KAZE uses AOS (Additive Operator Splitting) schemes since there are no analytical solution of the partial differential equations (PDEs) for nonlinear diffusion filtering. KAZE uses an adapted version of the SURF descriptor. Since the descriptors must work in a nonlinear scale-space model, the derivative responses are calculated and summed into a feature descriptor vector, and then the vector is centered at the feature. Finally, the descriptor is normalized into a unit vector.

AKAZE \cite{alcantarilla2011akaze} is the accelerated version of the KAZE algorithm. It benefits from FED (Fast Explicit Diffusion) in the detection step and uses Modified-Local Difference Binary (M-LDB) descriptor, which is a modified version of the original LDB descriptor \cite{yang2012ldb}, to create the feature descriptor vectors.

After features are detected and described with a feature extractor algorithm, they should be matched between frames. \textit{Mutual Nearest Neighbor Search (MNNS)} which is executed by measuring distances between descriptor vectors is commonly used for feature matching. The most common distance metrics are SAD (Sum of Absolute Differences), SSD (sum of squared differences), and Hamming distance. Moreover, while searching for nearest neighbors, Lowe \cite{lowe2004sift} proposed a method to reject ambiguous matches by thresholding the ratio of the distance of the closest match to the distance of the second closest one. This whole feature matching strategy is denoted as \textit{Mutual Nearest Neighbor Search with Bidirectional Ratio Test (MNNSwBRT)}.

\subsection{Learning-based Image Matching Algorithms}

Among the recently proposed learning-based techniques,  SuperPoint and SuperGlue algorithms follow the classical image matching pipeline, while Patch2Pix and DFM techniques directly output the matched features between two images.

SuperPoint \cite{detone2018superpoint} jointly detects keypoints and computes relevant descriptor vectors. In this method, the outputs of the MagicPoint \cite{detone2017magicpoint} detector is first exploited and a novel self-supervision strategy, namely Homographic Adaptation, is utilized to create a pseudo ground-truth interest point. Then, SuperPoint network is jointly trained in such a way that giving the keypoint confidence of each pixel and their corresponding descriptor vectors. 

SuperGlue \cite{sarlin2020superglue} can be considered as a feature matcher that computes the matches between two sets consisting of detected features and corresponding descriptor vectors. SuperGlue basically improves the descriptor vectors considering the cross and self attentions by the help of a graph neural network. At the  final step, SuperGlue algorithm learns to match features optimally by using differentiable Sinkhorn algorithm \cite{sinkhorn1967concerning, cuturi2013sinkhorn}.

Patch2Pix \cite{zhou2020patch2pix} starts by matching the deepest features of truncated ResNet-34 \cite{he2016resnet} network by employing NCNet \cite{rocco2020ncnet} as a matcher, and it continues refining feature locations until the pixel-level by utilizing mid and fine-level regressors which are weakly supervised by epipolar geometry constraints. 

The recent DFM method \cite{Efe_2021_CVPR} uses only a pre-trained classification network and well-studied conventional computer vision techniques, such as hierarchical refinement and ratio test. DFM first aligns two images by matching the terminal layers of the pre-trained VGG19 \cite{simonyan2014vgg} network; next, by the ratio test, DFM starts to match features of the terminal layers and refines those matches in a hierarchical way upto the first layers of VGG. 

It is an essential fact that all these learning-based approaches need training data. Specifically, SuperPoint exploits MagicPoint \cite{detone2017magicpoint} detector, which is trained using Synthetic Shapes dataset \cite{detone2017magicpoint}, and uses MS-COCO dataset \cite{lin2014mscoco} after labeling in a self-supervised manner. SuperGlue, on the other hand, is separately trained on different datasets, which are Oxford and Paris \cite{radenovic2018oxfordandparis}, ScanNet \cite{dai2017scannet}, and MegaDepth \cite{li2018megadepth}, for every particular problem, namely homography, indoor and outdoor. Patch2Pix utilizes ResNet34 \cite{he2016resnet} backbone, trained on ImageNet \cite{deng2009imagenet}, and its refinement network is trained on MegaDepth. Even DFM, which uses a pre-trained VGG-19 \cite{simonyan2014vgg} extractor, naturally needs this off-the-shelf network trained on ImageNet. Hence, all learning-based methods depend on the dataset characteristics, such as content and annotation quality. 

Finally, it should be noted that while Patch2Pix and DFM directly output the putative matches, SuperPoint requires a feature matcher at its final stage as classical algorithms; this matcher step might be either MNNSwBRT or SuperGlue.

\section{Experimental Setup}
\label{sec:experimentalsetup}
We have constructed an experimental setup in order to measure the performances of 5 classical and 4 learning-based image matching algorithms on HPatches dataset in terms of widely used metrics MMA and HEA by sliding only one parameter, either ratio test or confidence. 

\subsection{Dataset}
HPatches dataset \cite{balntas2017hpatches} consists of 116 sequences of two subsets, namely illumination and viewpoint sets. The illumination subset includes 57 sequences, and each has 6 images and 5 ground-truth homographies between the first image and others. The viewpoint subset has 59 sequences with the same structure. The sequences in the illumination subset have significant illumination variation with the same viewpoints antithetical to the viewpoint subset in which sequences have significant viewpoint changes with similar illuminations. Following D2-Net \cite{dusmanu2019d2net}, we left out large images and made evaluations on 52 illumination sequences and 56 viewpoint sequences in order to make all algorithms work and to be coherent with the literature. 

\subsection{Performance Metrics}

\subsubsection{Mean Matching Accuracy (MMA)}
Mean Matching Accuracy (MMA) is a widely used performance metric, and recently many state-of-the-art works \cite{dusmanu2019d2net, zhou2020patch2pix, parihar2021rord, Efe_2021_CVPR} reported their performances in terms of MMA on HPatches dataset. This metric basically measures the average accuracy of the matched features over the dataset. Given an image pair and matched features between them, matching accuracy is defined as the percentage of the correctly matched features. A match is accepted as a correct match if the distance between the reprojected feature point with ground-truth homography and its corresponding match point is less than given pixel threshold. In our experiments, we vary the threshold from 1 pixel to 10 pixels as in the literature.

\subsubsection{Homography Estimation Accuracy (HEA)}
Homography Estimation Accuracy (HEA) is another widely used metric for image matching evaluation and used in \cite{detone2018superpoint, zhou2020patch2pix, sun2021loftr, Efe_2021_CVPR} as a performance metric. MMA solely may not be sufficient for image matching evaluation, since an algorithm with a very limited number of and poorly distributed matches may make a high score in terms of MMA but it is likely to fail at geometric transformation estimation. Hence, we take into account HEA and use it as the second performance metric in all of our experiments. To compute HEA, four corners of one image is reprojected onto the other image with the estimated and the ground-truth homographies. Then we take the average distance between these projected points and accept the estimated homography correct, if the reprojection error is smaller than the given threshold. HEA is the rate of correctly estimated homographies over whole dataset.

\subsection{Algorithms}
We use OpenCV \cite{opencv_library} 4.5.2 implementations of the classical algorithms SIFT, ORB, KAZE, and AKAZE with the default parameters for feature detection and description. For SURF, we use OpenCV 3.4.2 with the default parameters. For SuperPoint, we utilize SuperGlue GitHub repository \cite{SuperGlueGithubRepo} in order to obtain keypoint locations and their descriptors. All the algorithms mentioned above perform Mutual Nearest Neighbor Search with Bidirectional Ratio Test (MNNSwBRT) to find matches between extracted features. We measure the performance for different ratio test thresholds from 0.1 to 1.0 with steps of 0.1. 

For DFM, we also used the original implementation \cite{DFMGithubRepo} of the algorithm which again takes the advantage of MNNSwBRT. However, since DFM benefits from MNNSwBRT multiple times, to be fair, we kept the ratio test thresholds for the deepest two layer’s descriptors fixed in the act of 0.95 and 0.90 as in the original paper, and only optimize the ratio test thresholds of the first three shallowest layer as their multiplication results with the threshold value. To illustrate, we have used the threshold set [0.80, 0.80, 0.80, 0.90, 0.95] for the threshold value 0.5 since $0.5^{(1/3)} \approx 0.80$.

For SuperGlue, we have used the GitHub repository \cite{SuperGlueGithubRepo} with the default parameters except for not resizing the input images and using the ‘outdoor’ setting, which gives better performance HPatches dataset. For Patch2Pix, we use the official GitHub repository \cite{Patch2PixGithubRepo} with default parameter settings. For both SuperGlue and Patch2Pix we measure the performance for different confidence thresholds from 0.9 to 0.0 with steps of 0.1.

\begin{table}[!t]
\vspace{0cm}
\centering
\adjustbox{max width=\columnwidth}{
  \begin{tabular}{l||c}
    \toprule
    {Method} & {Threshold} \\
    \midrule \\
    SIFT \cite{lowe2004sift} + NN & Ratio Test Threshold \\
    SURF \cite{bay2006surf} + NN & Ratio Test Threshold \\
    ORB \cite{rublee2011orb} + NN & Ratio Test Threshold \\
    KAZE \cite{alcantarilla2012kaze} + NN & Ratio Test Threshold \\
    AKAZE \cite{alcantarilla2011akaze} + NN & Ratio Test Threshold \\
    \midrule \\
    SuperPoint \cite{detone2018superpoint} + NN & Ratio Test Threshold \\
    SuperPoint + SuperGlue \cite{sarlin2020superglue} & (1 - Confidence) \\
    Patch2Pix \cite{zhou2020patch2pix} & (1 - Confidence) \\
    DFM \cite{Efe_2021_CVPR} & (Ratio Test Threshold)$^{3}$ for first 3 layers \\
    \bottomrule
  \end{tabular}
  }
  
\vspace{0.28cm}
\caption{\textbf{Definition of the Threshold Values} used in experiments. For all classical algorithms and SuperPoint + NN, we define \textit{threshold} as the \textit{ratio test threshold} of MNNSwBRT matcher, where for SuperGlue and Patch2Pix, it is defined as \textit{the complement of the confidence}. Finally, for DFM algorithm, the threshold is described as the \textit{product of the equal ratio test thresholds used for the first three layers}.}
\label{tab:table0}
\end{table}

Computing MMA is straightforward and reported with the same procedure in many works. Nonetheless, Homography Estimation performance depends on utilized homography estimation method, many image matching algorithms reported their results using different homography estimation methods. For example, Patch2Pix has used pydegensac, where DFM has used MATLAB’s estimateGeometricTransform function, both are different versions of RANSAC \cite{fischler1981ransac} algorithm. In this work, we use OpenCV’s findHomography function with newly introduced cv.USAC\_MAGSAC method \cite{barath2020usacmagsac}, which is available from OpenCV 4.5.2 version, and whose success is demonstrated in \cite{USAC_MAGSAC}. We adjust the other RANSAC’s parameters as following: ransacReprojThreshold=3.0, maxIters=5000, confidence=0.9999.

\section{Experimental Results}
We followed the procedure explained in Section \ref{sec:experimentalsetup} and evaluated nine algorithms on HPatches dataset, with varying thresholds, which are different for each algorithm and defined in Table \ref{tab:table0}.

\subsection{Effect of Matching Threshold}

\begin{figure*}[!ht]
\vspace{0cm}
\centering
\includegraphics[width=0.93\textwidth]{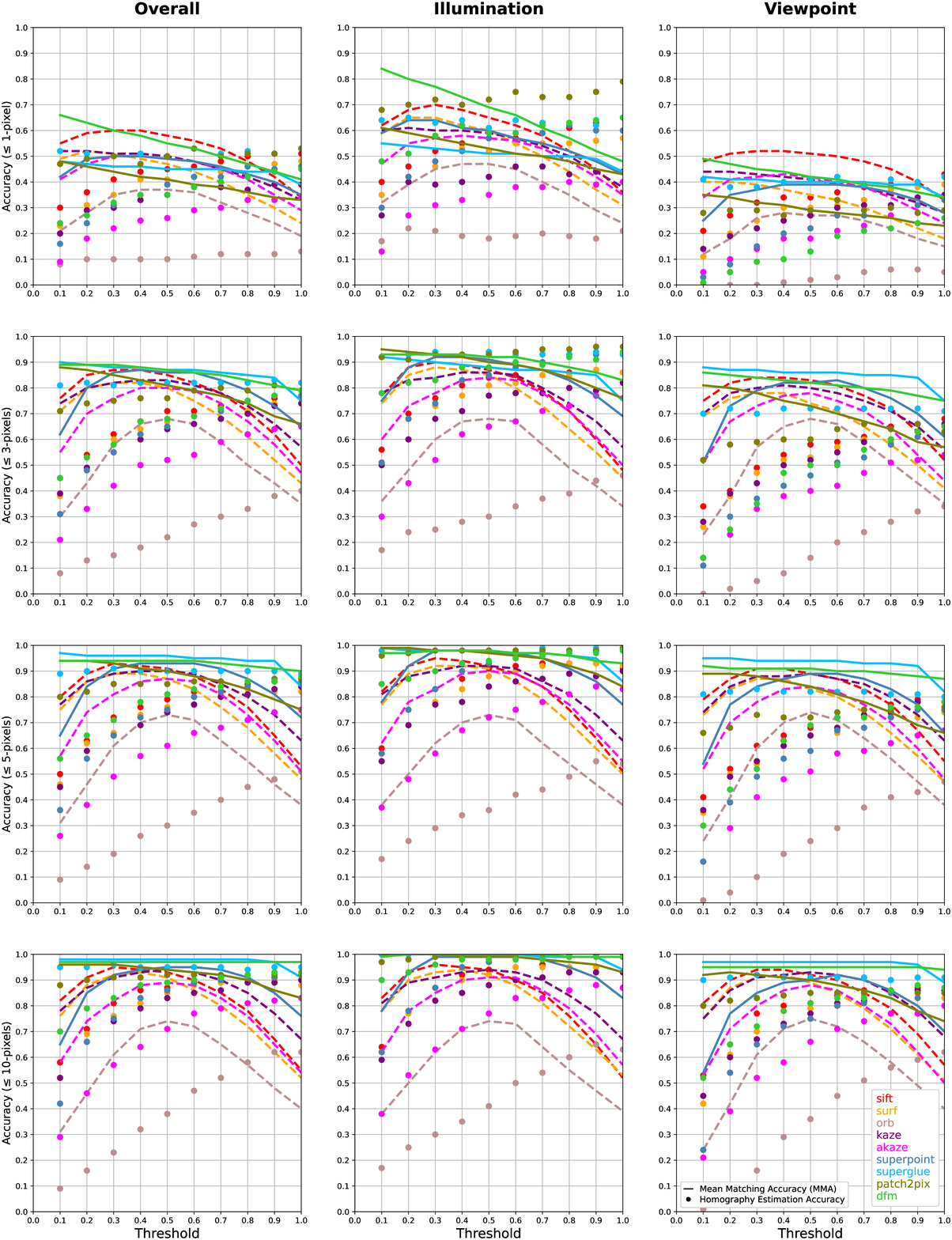}
\smallbreak
\caption{\textbf{Experimental Results} of 5 classical and 4 learning-based algorithms in terms of MMA and HEA on HPatches \cite{balntas2017hpatches} dataset. MMA results are shown as dashed lines for classical algorithms and straight lines for learning-based algorithms, where HEA results are shown as dots. Individual plots show the accuracy with varying threshold values while rows indicate different pixel thresholds.}

\label{fig:figure1}
\end{figure*}

\begin{table*}[btp]
\vspace{0cm}
\adjustbox{max width=\textwidth}{
  \begin{tabular}{l||cccccccc||cccccccc}
    \toprule
      \multirow{4}{*}{Method} &
      \multicolumn{8}{c}{Mean Matching Accuracy (MMA)} &
      \multicolumn{8}{c}{Homography Estimation Accuracy (HEA)}\\
   
    {} & \multicolumn{2}{c}{$\leq$1px} & \multicolumn{2}{c}{$\leq$3px} & \multicolumn{2}{c}{$\leq$5px} & \multicolumn{2}{c}{$\leq$10px} & \multicolumn{2}{c}{$\leq$1px} & \multicolumn{2}{c}{$\leq$3px} & \multicolumn{2}{c}{$\leq$5px} & \multicolumn{2}{c}{$\leq$10px} \\
    \cmidrule{2-17} \\
    {} & {Acc.} & {Thr.} & {Acc.} & {Thr.} & {Acc.} & {Thr.} & {Acc.} & {Thr.} & {Acc.} & {Thr.} & {Acc.} & {Thr.} & {Acc.} & {Thr.} & {Acc.} & {Thr.}\\
    \midrule \\
    SIFT \cite{lowe2004sift} + NN & \textbf{\textit{0.60}} & 0.3 & 0.87 & 0.3 & 0.93 & 0.3 & 0.95 & 0.3 & \textbf{\textit{0.51}} & 0.9 & 0.76 & 0.9 & 0.86 & 0.9 & 0.91 & 0.7  \\
    SURF \cite{bay2006surf} + NN & 0.52 & 0.2 & 0.82 & 0.3 & 0.89 & 0.3 & 0.93 & 0.4 & 0.48 & 1.0 & 0.74 & 0.9 & 0.83 & 0.8 & 0.91 & 0.8  \\
    ORB \cite{rublee2011orb} + NN & 0.37 & 0.4 & 0.68 & 0.5 & 0.73 & 0.5 & 0.74 & 0.5 & 0.13 & 1.0 & 0.4 & 1.0 & 0.51 & 1.0 & 0.62 & 0.9 \\
    KAZE \cite{alcantarilla2012kaze} + NN & 0.52 & 0.1 & 0.83 & 0.4 & 0.90 & 0.4 & 0.94 & 0.5 & 0.39 & 0.9 & 0.74 & 1.0 & 0.84 & 1.0 & 0.90 & 1.0  \\
    AKAZE \cite{alcantarilla2011akaze} + NN & 0.50 & 0.3 & 0.80 & 0.4 & 0.87 & 0.5 & 0.89 & 0.5 & 0.34 & 1.0 & 0.65 & 1.0 & 0.75 & 1.0 & 0.83 & 1.0  \\
    \midrule \\
    SuperPoint \cite{detone2018superpoint} + NN & 0.50 & 0.3 & 0.87 & 0.4 & 0.93 & 0.4 & 0.95 & 0.5 & 0.46 & 0.8 & 0.70 & 0.9 & 0.86 & 0.9 & 0.92 & 0.8  \\
    SuperPoint + SuperGlue \cite{sarlin2020superglue} & 0.48 & 0.1 & \textbf{0.90} & 0.1 & \textbf{0.97} & 0.1 & \textbf{0.98} & 0.1 & \textbf{0.53} & 0.6 & \textbf{0.83} & 0.9 & \textbf{0.91} & 0.3 &  \textbf{0.96} & 0.3\\
    Patch2Pix \cite{zhou2020patch2pix} & 0.48 & 0.1 & 0.88 & 0.1 & \textbf{\textit{0.94}} & 0.1 & 0.96 & 0.1 & \textbf{0.53} & 0.6 & \textbf{\textit{0.81}} & 0.9 & 0.87 & 0.9 & 0.92 & 0.5  \\
    DFM \cite{Efe_2021_CVPR} & \textbf{0.66} & 0.1 & \textbf{\textit{0.89}} & 0.1 & \textbf{\textit{0.94}} & 0.1 & \textbf{\textit{0.97}} & 0.1 & 0.45 & 1.0 & 0.79 & 1.0 & \textbf{\textit{0.88}} & 1.0 & \textbf{\textit{0.93}} & 0.9 \\
    \bottomrule
  \end{tabular}
  }
  
\vspace{0.28cm}
\caption{\textbf{Best accuracy values} and their relevant thresholds for each evaluated algorithm in terms of both MMA and HEA metrics, considering the 1, 3, 5, and 10-pixel thresholds. Best-performing results are shown as bold and the second-best-performing ones are shown as bold-italic.}
\label{tab:table1}
\end{table*}

Figure \ref{fig:figure1} illustrates the performances of 5 classical and 4 learning-based image matching algorithms with varying thresholds in terms of MMA and HEA with different pixel thresholds on HPatches dataset. The threshold is defined as the ratio test threshold for MNNSwBRT matcher for SIFT, SURF, ORB, KAZE, AKAZE, and SuperPoint algorithms, (ratio test threshold)$^{3}$ for DFM and (1 – confidence) for SuperGlue and Patch2Pix algorithms. A smaller threshold means strict matching and returns fewer matches, while larger threshold values mean loose matching and return more matches.

In Figure \ref{fig:figure1}, it can be clearly observed that the threshold has a notable effect on the performance. For example, for a specific threshold (0.4), SIFT becomes the best-performing method, while it takes 4th place, when the threshold value is 1.0 in terms of MMA under 1 pixel threshold. A similar pattern is observed for most of the algorithms evaluated. Another interesting observation is that most of the classical algorithms have a bell-shaped curve, similar to the observations in \cite{jin2021imagematchingfrompapertopractice}. Moreover, the effect of the threshold is more significant than learning-based algorithms in terms of MMA, which claims that when evaluating a classical method, at least the ratio test threshold parameter should be optimized for the given dataset. An additional important observation is the almost monotonic increase in HEA for every algorithm, meaning that RANSAC is powerful enough to handle putative match sets consists of some amount of outliers and further giving better results.

\subsection{Comparisons based on Best Accuracy Values}

Table \ref{tab:table1} is established by picking the best accuracy in terms of both MMA and HEA and the selected thresholds for each algorithm, considering the 1, 3, 5, and 10-pixel threshold. From the table, we again notice the monotonic increasing behavior of HEA with respect to threshold, and in contrast, MMA maximizes in lower threshold values. That means strict thresholds result in correct matches while loose thresholds result in more matches and increase the accuracy of the homography estimation with the help of RANSAC. Last but not least, the implication from the table is that there is only a small performance gap between recently developed state-of-the-art learning-based algorithms and the classical algorithms with only adjusting a single parameter. For example, SIFT is the second-best performing algorithm under 1-pixel accuracy, and it has only a few percent away from the best-performing methods for the other pixel thresholds. Noting that we did not attempt to optimize any other parameters of the classical algorithms, such an optimization over the dataset might make these algorithms outperform the state-of-the-art.

\subsection{Comparisons based on Best Area Under Accuracy Curves}

\begin{table}[b]
\vspace{0cm}
\adjustbox{max width=\columnwidth}{
  \begin{tabular}{l||c||ccc||ccc}
    \toprule
      \multirow{2}{*}{Method} &
      \multicolumn{1}{c}{} &
      \multicolumn{3}{c}{MMA} &
      \multicolumn{3}{c}{HEA}\\
    {} & {\#Features} & {AUC} & {Thr.} & {\#Matches} & {AUC} & {Thr.} & {\#Matches} \\
    \midrule \\
    SIFT \cite{lowe2004sift} + NN & 4572 & \textbf{\textit{89.6}} & 0.3 & 478 & 82.1 & 0.9 & 1293  \\
    SURF \cite{bay2006surf} + NN & 6003 & 85.6 & 0.3 & 276 & 80.0 & 0.9 & 1378 \\
    ORB \cite{rublee2011orb} + NN & 499 & 69.1 & 0.5 & 19 & 48.0 & 1.0 & 170  \\
    KAZE \cite{alcantarilla2012kaze} + NN & 3120 & 86.3 & 0.5 & 544 & 79.8 & 1.0 & 1287  \\
    AKAZE \cite{alcantarilla2011akaze} + NN & 2694 & 82.9 &  0.5 & 219 & 72.0 & 1.0 & 1011\\
    \midrule \\
    SuperPoint \cite{detone2018superpoint} + NN & 921 & 88.3 & 0.5 & 253 & 82.6 & 1.0 & 506  \\
    SuperPoint + SuperGlue \cite{sarlin2020superglue} & 921 & \textbf{91.6} & 0.1 & 441 & \textbf{86.8} & 0.6 & 482 \\
    Patch2Pix \cite{zhou2020patch2pix} & - & 89.2 & 0.1 & 723 & \textbf{\textit{84.1}} & 0.9 & 1434 \\
    DFM \cite{Efe_2021_CVPR} & - & \textbf{91.6} & 0.1 & 881 & 84.0 & 1.0 & 16619 \\
    \bottomrule
  \end{tabular}
  }

\vspace{0.28cm}
\caption{\textbf{Best area under curve (AUC) percentage values} and their relevant thresholds for each algorithm considering both MMA and HEA metrics from 1 to 10 pixel threshold. Best-performing results are shown as bold, and the second-best-performing ones are shown as bold-italic. Also, the number of detected features and the number of matched features using related threshold values are indicated.}
\label{tab:table2}
\end{table}

\begin{figure}[t]
\vspace{0cm}
\includegraphics[width=1.0\columnwidth]{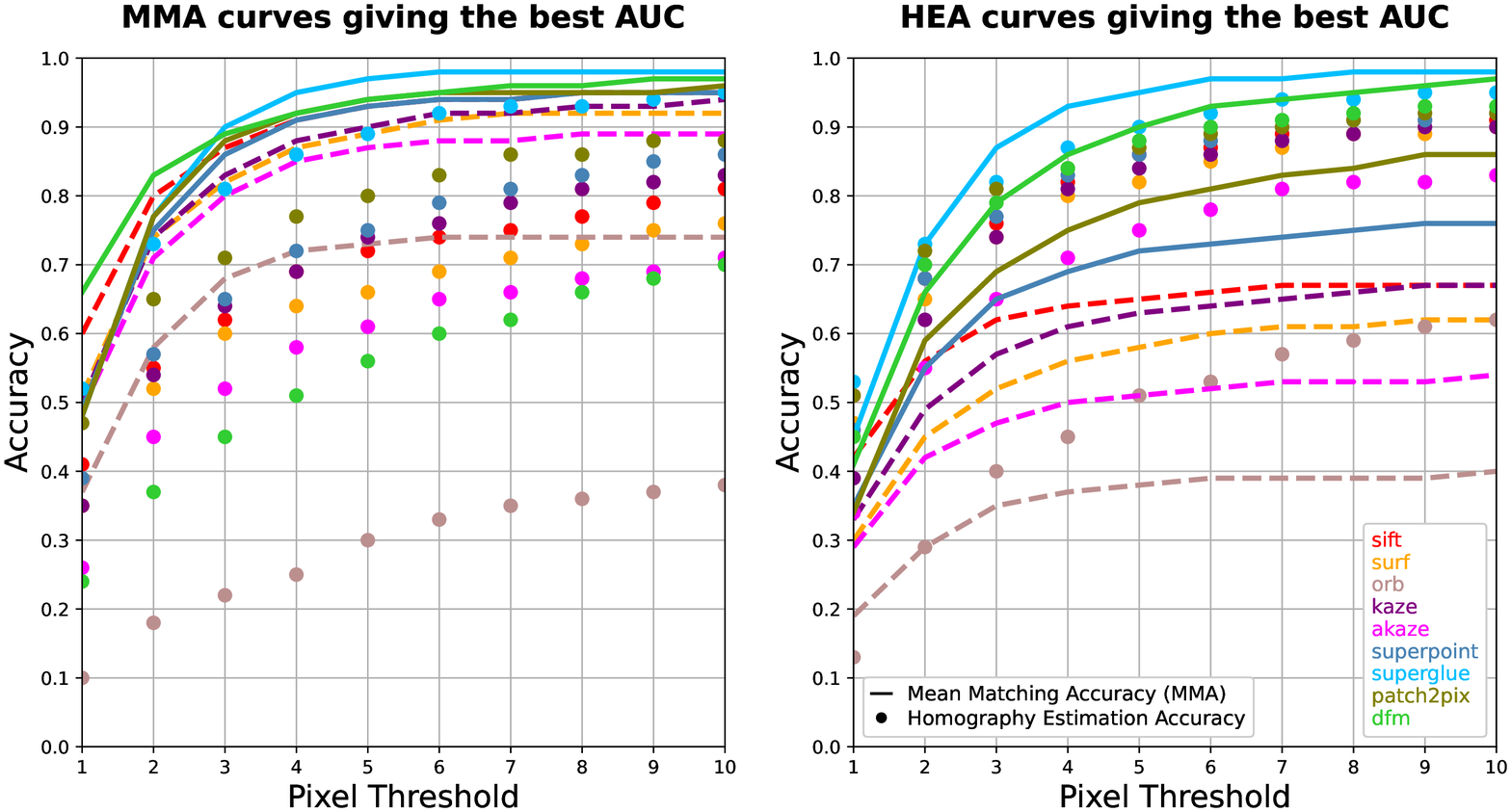}
\smallbreak
\caption{\textbf{Overall MMA and HEA curves} that give the best AUC for each algorithm. These curves indicate the best possible AUC result for each algorithm can achieve on HPatches \cite{balntas2017hpatches} dataset by only optimizing a single parameter ratio test or match confidence. For each setting that gives the best AUC for one metric, the result of the other metric is also presented.}
\label{fig:figure2}
\end{figure}

Table \ref{tab:table2} is constructed by selecting a threshold for each algorithm to maximize the average accuracy for different pixel thresholds from 1 to 10 pixels. This average gives area under the accuracy curve (AUC) shown in Figure \ref{fig:figure2}. Table \ref{tab:table2} also illustrates the number of matches for relevant thresholds, indicating the fact that high MMA can be achieved with higher confidence and less number of matches, while high HEA can be obtained with loose confidence and hence a vast number of matches. The most cardinal demonstration of Table \ref{tab:table2} is a classical algorithm SIFT is the second best performing algorithm in terms of AUC of mean matching accuracy. Also, its performance in terms of AUC of homography estimation accuracy is only two percentage below the second best performing algorithm, Patch2Pix. In addition, SURF and KAZE algorithms also have comparable performances with the state-of-the-art algorithms. Finally, note that although we categorize DFM as a learning-based method, it can be considered as almost a classical approach as it utilizes the deep features extracted by a pre-trained VGG-19 network, only employs the well established classical computer vision algorithms such as initial warping, hierarchical refinement and nearest neighbor search with ratio test in a very simple framework; and underline that DFM has no specific training procedure for image matching task. 

Figure \ref{fig:figure2} exhibits the MMA and HEA curves that give the best AUC for each algorithm. The figure utilizes the thresholds given in Table \ref{tab:table2} and illustrates the MMA and HEA curves. Note that it also shows the curve of the other metric by using the thresholds optimized for one metric, i.e., it displays the HEA curve with the thresholds that maximize the AUC of the MMA curve and vice-versa. It can be seen from the figure, for the most of the algorithms, the best threshold for one metric is not good enough for the other metric. For example, when DFM's ratio test threshold is optimized for MMA, its performance is poor for HEA. Except that, Patch2Pix and specifically SuperGlue algorithms are robust to such a parameter variation, performing well in both metrics. It may result from the fact that these two algorithms inherently learn the scene geometry so that optimizing these algorithms to obtain high MMA will naturally result in a high performance for HEA as well. Most importantly, the figure claims that classical methods can still perform very close to the well-trained state-of-the-art algorithms by adjusting just a single parameter. In fact, SIFT still achieves state-of-the-art performance, being under only DFM in terms of MMA for 1 and 2-pixel thresholds, and there is not a significant performance gap between the state-of-the-art algorithms in terms of HEA. This kind of result, meaningly a classical algorithm performs this much closer to the recent methods, is not reported in previous studies \cite{dusmanu2019d2net, zhou2020patch2pix, parihar2021rord, Efe_2021_CVPR}.

\section{Conclusions}
In this study, we have evaluated five classical and four learning-based algorithms for image matching task on well-known HPatches dataset \cite{balntas2017hpatches}. We demonstrated the effect of the ratio test threshold for feature matching through experiments. Furthermore, we showed that classical methods are quite powerful so that they still perform very close to state-of-the-art. Specifically, one of the most popular methods nearly two decades, SIFT \cite{lowe2004sift} achieves almost state-of-the-art performance by only optimizing ratio test threshold, which is neglected  in previous studies. DFM \cite{Efe_2021_CVPR} algorithm also demonstrates the feature matching capability of classical techniques by achieving state-of-the-art performance with only practicing well-established classical computer vision techniques on top of a pre-trained deep learning backbone.

Although our arguments may be limited with HPatches dataset, we argue that classical methods should be carefully analyzed before immediately working on a learning-based method. Despite the promising performance of the traditional methods, we emphasize a learning-based technique, SuperPoint \cite{detone2018superpoint} + SuperGlue \cite{sarlin2020superglue}, as the best-performing method on HPatches dataset in terms of area under curves for mean matching accuracy and homography estimation accuracy along with its robustness to the varying confidence thresholds.  

{\small
\bibliographystyle{ieee}
\bibliography{egbib}
}

\end{document}